\pdfoutput=1

\documentclass[times,final]{elsarticle}
\usepackage[utf8]{inputenc}
\usepackage{url}
\usepackage{xcolor}
\usepackage{todonotes}

\definecolor{myblack}{rgb}{0.0, 0.0, 0.0}

\newcommand{\virgolette}[1]{``#1''}

\journal{
Future Generation Computer Systems
}

\begin{document}

\begin{frontmatter}
\title{Generating Knowledge Graphs by Employing Natural Language Processing and Machine Learning Techniques within the Scholarly Domain}

\author[1,2,3]{Danilo Dess\`{i}\corref{cor1}} \cortext[cor1]{Corresponding author. Tel. +39-070-675-8756.} \ead{danilo\_dessi@unica.it}
\author[4]{Francesco Osborne}
\author[1]{Diego Reforgiato Recupero}
\author[5]{Davide Buscaldi}
\author[4]{Enrico Motta} 

\address[1]{Department of Mathematics and Computer Science, University of Cagliari, Cagliari, Italy}
\address[2]{FIZ Karlsruhe - Leibniz Institute for Information Infrastructure, Germany}
\address[3]{Karlsruhe Institute of Technology, Institute AIFB, Germany}
\address[4]{Knowledge Media Institute, The Open University, Milton Keynes, UK}
\address[5]{LIPN, CNRS (UMR 7030), University Paris 13, Villetaneuse, France}

\begin{abstract}
The continuous growth of scientific literature brings innovations and, at the same time, raises new challenges. One of them is related to the fact that its analysis has become difficult due to the high volume of published papers for which manual effort for annotations and management is required. Novel technological infrastructures are needed to help researchers, research policy makers, and companies to time-efficiently browse, analyse, and forecast scientific research. 
Knowledge graphs i.e., large networks of entities and relationships, have proved to be effective solution in this space. Scientific knowledge graphs focus on the scholarly domain and typically contain metadata describing research publications such as authors, venues, organizations, research topics, and citations. However, the current generation of knowledge graphs lacks of an explicit representation of the knowledge presented in the research papers. As such, in this paper, we present a new architecture that takes advantage of Natural Language Processing and Machine Learning methods for extracting entities and relationships from research publications and integrates them in a large-scale knowledge graph. 
Within this research work, we i) tackle the challenge of knowledge extraction by employing several state-of-the-art Natural Language Processing and Text Mining tools, ii) describe an approach for integrating entities and relationships generated by these tools, iii) show the advantage of such an hybrid system over alternative approaches, and vi) as a chosen use case, we generated a scientific knowledge graph including $109,105$ triples, extracted from $26,827$ abstracts of papers within the \textit{Semantic Web} domain. As our approach is general and can be applied to any domain, we expect that it can facilitate the management, analysis, dissemination, and processing of scientific knowledge.
\end{abstract}
\end{frontmatter}

\section{Introduction}\label{intro}


Nowadays, we are seeing a constant growth of scholarly knowledge, making the access to scholarly contents more and more challenging through traditional search methods. This problem has been partially solved thanks to digital libraries which provide scientists with tools to explore research papers and to monitor research topics. Nevertheless, the dissemination of scientific information is mainly document-based and mining contents requires human manual intervention, thus limiting chances to spread knowledge and its automatic processing~\cite{jaradeh2019open}.

Despite the large number and variety of tools and services available today for exploring scholarly data, current support is still very limited in the context of sensemaking tasks that require a comprehensive and accurate representation of the entities within a domain and their semantic relationships. 
This raises the need of more flexible and fine-grained scholarly data representations that can be used within technological infrastructures for the production of insights and knowledge out of the data~\cite{buscaldi2019mining,tennant2019ten,buscaldi2019poster}. 
Kitano~\cite{kitano2016artificial} proposed a similar and more ambitious vision, suggesting the development of an artificial intelligence system able to make major scientific discoveries in biomedical sciences and win a Nobel Prize.

Among the existing representations, knowledge graphs i.e., large networks of entities and relationships, usually expressed as RDF triples, relevant to a specific domain or an organization~\cite{ehrlinger2016towards},  provide a great method to organize information in a structured way. They already have been successfully used to understand complex processes in various domains such as social networks ego-nets~\cite{lan2016learning} and biological functions~\cite{dessi2018supernoder}.

Tasks like question answering, summarization, and decision support have already benefited from these structured representations. The generation of knowledge graphs from unstructured source of data is today key for data science and researchers across various disciplines (e.g., Natural Language Processing (NLP), Information Extraction, Machine Learning, and so on.) have been mobilized to design and implement methodologies to build them.  State-of-the-art projects such as DBPedia~\cite{lehmann2015dbpedia}, Google Knowledge Graph, BabelNet\footnote{\url{https://babelnet.org/}}, and YAGO\footnote{\url{https://www.mpi-inf.mpg.de/departments/databases-and-information-systems/research/yago-naga/yago/}} build Knowledge Graphs by harvesting entities and relations from textual resources (e.g., Wikipedia pages).  The creation of such knowledge graphs is a complex process that typically requires the extraction and integration of various information from structured and unstructured sources.

Scientific knowledge graphs focus on the scholarly domain and typically contain metadata describing research publications such as  authors, venues, organizations, research topics, and citations. Some examples are Open Academic Graph\footnote{\url{https://www.openacademic.ai/oag/}}, Scholarlydata.org~\cite{nuzzolese2016conference}, Microsoft Academic Graph\footnote{\url{https://www.microsoft.com/en-us/research/project/microsoft-academic-graph}}~\cite{wang2020microsoft} (MAG), Scopus\footnote{\url{https://www.scopus.com/}}, Semantic Scholar\footnote{\url{https://www.semanticscholar.org/}}, Aminer~\cite{zhang2018name}, Core~\cite{knoth2012core}, OpenCitations~\cite{peroni2020opencitations}, and Dimensions\footnote{\url{https://www.dimensions.ai/}}. 
These resources provide substantial benefits to researchers, companies, and policy makers by powering data-driven services for navigating, analyzing, and making sense of research dynamics.
However, the current generation of knowledge graphs lacks of an explicit representation of research knowledge discussed in the scientific papers. This is usually only described by not machine-readable metadata, such as natural language text in the title and abstract, and in some cases a  list of topics or keywords from a domain vocabulary or taxonomy (e.g., MeSH\footnote{\url{https://www.nlm.nih.gov/mesh/meshhome.html}}, ACM Digital Library\footnote{\url{https://dl.acm.org/}}, PhySH\footnote{\url{https://physh.aps.org/}}, CSO\footnote{\url{https://cso.kmi.open.ac.uk/home}}). These data  are useful to some degree, but do not offer a formal description of the nature and the relationships of relevant research entities. For instance
this representation does not give us any information about what "sentiment analysis" is and how it interlinks with other entities in the research domain. 
It would be much more useful to know that this is a sub-task of Natural Language Processing that aims at detecting the polarity of users opinion by applying a range of machine learning approaches on reviews and social media data such as twitter posts.

A robust and formal representation of the content of scientific publications that types and interlinks research entities would enable many advanced tasks that are not supported by the current generation of systems. For instance, it would allow to formulate complex semantic queries about research knowledge such as "return all approaches and benchmarks that are used to detect fake news". It would also support tools for the exploration of research knowledge by allowing users to navigate the different semantic links and retrieve all publications associated with specific claims. It could also enable a new generation of academic recommendation systems and tools for hypothesis generation.

The Semantic Web community has been working for a while on the generation of machine-readable representations of research, by fostering the Semantic Publishing paradigm ~\cite{shotton2009semantic}, creating bibliographic repositories in the Linked Data Cloud ~\cite{nuzzolese2016semantic}, generating knowledge bases of biological data ~\cite{belleau2008bio2rdf}, formalising research workflows ~\cite{wolstencroft2013taverna}, implementing systems for managing nano-publications~\cite{groth2010ana,kuhn2016decentralized} and  micropublications ~\cite{schneider2014using}, and developing a variety of ontologies to describe scholarly data, e.g., SWRC \footnote{SWRC - \url{http://ontoware.org/swrc}}, BIBO \footnote{BIBO - \url{http://bibliontology.com}}, BiDO\footnote{BiDO - \url{http://purl.org/spar/bido}}, FABIO\footnote{FABIO - \url{http://purl.org/spar/fabio}}, SPAR\footnote{SPAR - \url{http://www.sparontologies.net/}\cite{peroni2018spar}}, CSO\footnote{CSO - \url{https://cso.kmi.open.ac.uk/}~\cite{salatino2019b}}, and SKGO\footnote{SKGO - \url{https://github.com/saidfathalla/Science-knowledge-graph-ontologies}}~\cite{fathalla2020towards}.  
Some recent solutions, such as RASH\footnote{RASH - \url{https://github.com/essepuntato/rash}}~\cite{peroni2017research}, and the Open Research Knowledge Graph\footnote{ORKG - \url{https://www.orkg.org/orkg/}}~\cite{auer2018towards} highlighted the advantages of describing research papers in a structured manner.
However, the resulting knowledge bases still need to be manually populated by domain experts, which is a time consuming and expensive process. We still lack systems able to extract knowledge from large collection of research publications and automatically generate a comprehensive representation of research concepts.

It follows that a significant open challenge in this domain regards the automatic generation of scientific knowledge graphs that  contain an explicit representation of the knowledge presented in scientific publications~\cite{auer2018towards}, and  describe entities such as approaches, claims, applications, data, results reported in each paper. The resulting knowledge base would be able to support a new generation of content-aware services for exploring the research environment at a much more granular level. 

Most of the relevant information for populating such a knowledge graph might be derived from existing textual elements of research publications. To such an aim, in the last years, we assisted to the emergence of several excellent Machine Learning and NLP tools for entity linking and relationship extraction~ \cite{mesbah2018tse,auer2018towards,GangemiEtAl2017,martinez2018openie,luan2018multi}. However, integrating the output of these tools in a coherent and comprehensive knowledge graph is still an open issue.

For instance, different tools may use different lexical resources, named-entity recognition approaches, and training sets and thus will often label the same entities with different names and disagree on the relation between them. 


In this paper, we present a novel architecture that uses an ensemble of NLP and Machine Learning methods for extracting entities and relationships in form of triples from research publications, and then integrates them in a knowledge graph using Semantic Web best practices.
The main hypothesis behind this work is that an hybrid framework combining both supervised and unsupervised methods will produce the most comprehensive set of triples (i.e., high recall) while still yielding a good precision.


Within our work, we refer to an entity as a statement that indicates an object (e.g., a topic, a tool name, a well-known algorithm, etc.). We create a relation between two entities when they are syntactically or semantically connected. As an example, if a tool \textit{T} employs an algorithm \textit{A}, we may build the triple \textit{\textless \texttt{T}, \texttt{employ},  \texttt{A} \textgreater}.
We compared our approach versus alternative methods on a manually annotated gold standard covering the Semantic Web domain.

The main contributions of the research presented in this paper are therefore the following:
\begin{itemize}
    \item we propose an architecture that combines various tools for extracting entities and relations from research publications;
    \item we employ Semantic Web best practices, statistics, NLP, and Machine Learning techniques for integrating these entities and triples;
    \item we show the advantage of an hybrid approach versus methods that are only focused on supervised classification (e.g., Luan Yi et al. in~\cite{luan2018multi})  or NLP tools (e.g., OpenIE);
    \item we carry out an evaluation of the resulting triples in terms of precision, recall, and F-measure;
    \item we generated a gold standard of manually annotated triples that can be used as benchmark for this task.
\end{itemize}{}

In this paper we focus on the Semantic Web as main domain, but the resulting approach is general and can be applied to any other domain. 
The code of the framework, the extracted triples, and the gold standard used in the evaluation are available through a GitHub repository\footnote{\url{https://github.com/danilo-dessi/skg}}. 

The remainder of this paper is organized as follows.  Section~\ref{scholarly-problem} formalizes the problem we addressed. The proposed methodology is detailed in Section~\ref{scholarly-methodology}. The evaluation and its discussion are reported in Section~\ref{scholarly-result}.  Section~\ref{related} discusses the related work and highlights the main differences with the proposed approach. Finally, Section~\ref{conclusion} concludes the paper, explains limitations that still exist, and defines future research works.

\section{Problem Statement}\label{scholarly-problem}


Given a large collection of research papers, we want to generate a large-scale knowledge base, that will include all relevant entities in a certain domain and their relationships.

More in detail, given a set of scientific documents $D = \{d_1, \ldots, d_n\}$, we build a model $\gamma : D \rightarrow T$, where $T$ is a set of triples (also referred as relationships) $(s,p,o)$ where $s$ and $o$ belong to a set of entities $E$ and $p$ belongs to a set of relations labels $L$. Each triple needs also to be associated with the set of papers it was extracted from, allowing to assess 
how the claim is supported  in the original collection of documents. 

The resulting knowledge graph can be employed for different problems of new research fields (e.g., detection of research communities, their dynamics and trends, forecasting of research dynamics using sentiment analysis, measuring fairness of open access datasets, etc.), and, in general, as a support resource for scientists in conducting scientific research.

\section{Methodology}\label{scholarly-methodology}
\begin{figure}
    \centering
    \includegraphics[width=1.0\linewidth]{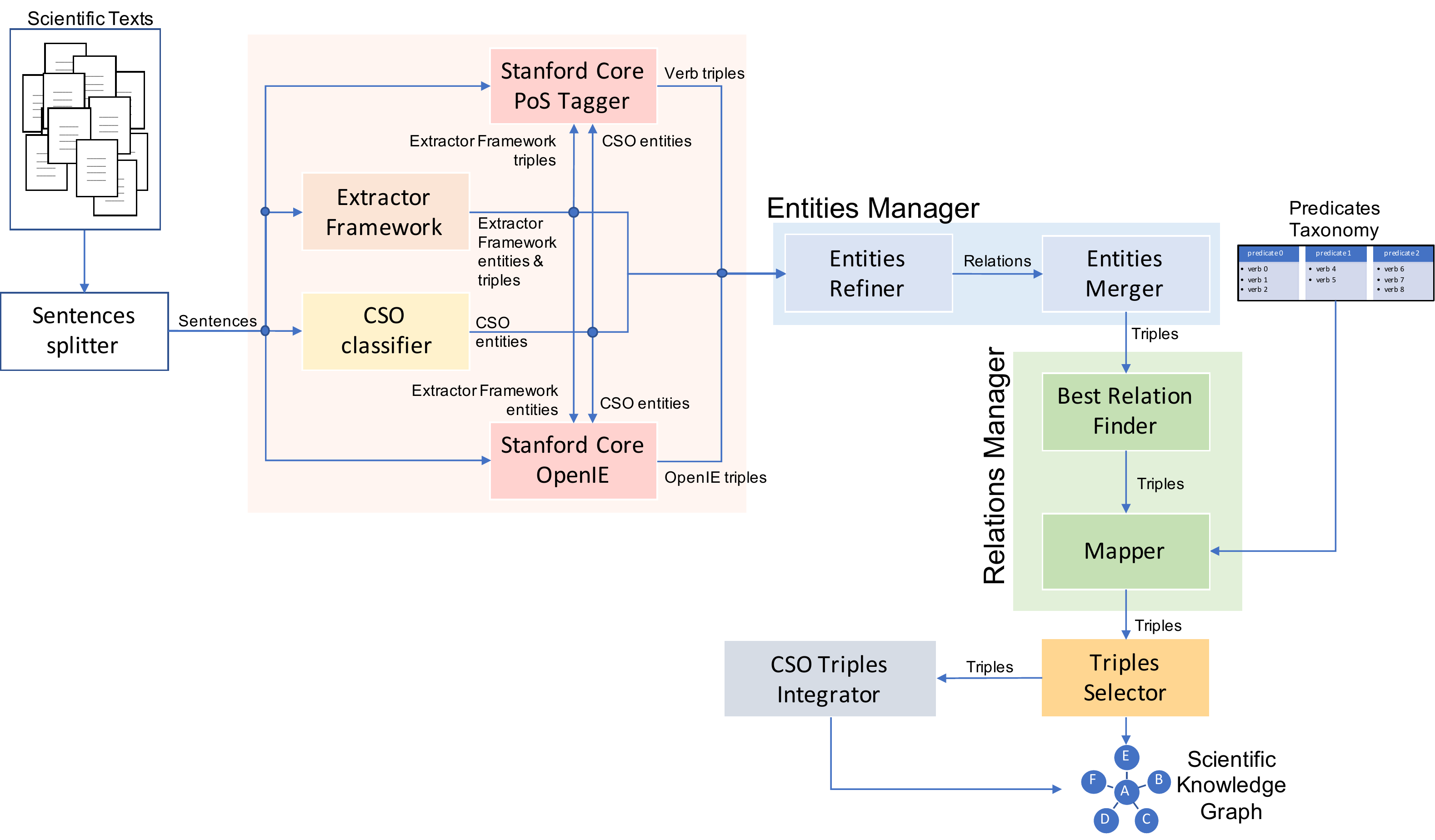}
    \caption{Workflow of our approach for building a scientific knowledge graph from scientific textual resources.}
    \label{fig:Scientific Knowledge Graph_schema}
\end{figure}

In this section, we describe the approach that we applied to produce a scientific knowledge graph of research entities. The workflow of our pipeline is shown in Figure~\ref{fig:Scientific Knowledge Graph_schema}.
In short, our framework includes the following steps:
\begin{enumerate}
  \item \textbf{Extraction of entities and triples}, which exploits an ensemble of several NLP and machine learning tools to extract triples from text.
  \item \textbf{Entity refining}, in which the resulting entities are merged and cleaned up.
  \item \textbf{Triple refining}, in which the triples extracted by the different tools are merged together and the relations are mapped to a common vocabulary.
   \item \textbf{Triple selection}, in which we select the set of "trusted"  triples that will be included in the output by first creating a smaller knowledge graph composed by triples associated with a good number of papers and then enriching this set with other semantically consistent triples. 
In the following subsection we will describe the architecture in more details and discuss the specific NLP and machine learning tools that we used in the implementation of our prototype. 

\end{enumerate}


\subsection{Extraction of Entities and Relations}
\label{extraction_entities_relations}
For extracting entities and relations, we exploited the following methods:
\begin{itemize}
    \item \emph{The extractor framework~\cite{luan2018multi}} designed by Luan Yi et al. that we modified and embedded within our pipeline. It is based on Deep Learning models and provides modules for detecting entities and relations from scientific literature. It detects six types of entities (\textit{Task, Method, Metric, Material, Other-Scientific-Term, and Generic}) and seven types of relations among a list of predefined choices (\textit{Compare, Part-of, Conjunction, Evaluate-for, Feature-of, Used-for, Hyponym-Of}). For the purpose of this work, we discarded all the triples with relation \textit{Conjunction}, since they were too generic. In particular, the extractor framework uses feed-forward neural networks over span representations of the input texts to compute two scores $v_1$ and $v_2$. The score $v_1$ is computed on single spans and measures how likely a span may be associated to an entity type. The second score $v_2$ is a pairwise score on a pair of span representations and measures how likely spans are involved in a relation. Therefore, for a given pair of span representations, let's say $(t_1, t_2)$, the scores $v^{t_1}_1$, $v^{t_2}_1$, and $v^{(t_1, t_2)}_2$ are computed. If both $v^{t_1}_1$ and $v^{t_2}_1$ meet a threshold $t_{entity}$, and $v^{(t_1, t_2)}_2$ meets a threshold $t_{relation}$ then the span representations $t_1$ and $t_2$ are labelled as entities, and their pair as relationship $(t_1, t_2)$. The type of entity where the value $v_1$ is the highest is associated to the entity itself. Similarly, a pair $(t_1, t_2)$ is associated to the type of relation $r$ where the pair has the highest value of $v_2$, yielding the triple $(t_1, r, t_2)$. For example, from the following sentence \virgolette{\textit{We propose a new web recommendation system based on reinforcement learning.}}, this framework detected \textit{web recommendation system} as a \textit{Task}, \textit{reinforcement learning} as a \textit{Method}, and the relation \textit{Used-for} between them, yielding the triple \textit{\textless reinforcement learning, Used-for,	web recommendation system \textgreater}. 
    We refer to this framework as Extractor Framework. 
    
    \item \emph{The CSO Classifier~\cite{salatino2019cso}}\footnote{\url{https://github.com/angelosalatino/cso-classifier}}, a tool for automatically classifying research papers according to the Computer Science Ontology (CSO)\footnote{\url{http://cso.kmi.open.ac.uk}}~\cite{salatino2018computer}, which is a comprehensive automatically generated ontology of research areas in the field of Computer Science.
    The current version of CSO describes 14K research topics arranged in a nine level polyhierarchical taxonomy.
    The CSO classifier identifies topics by means of two different components, the syntactic module and the semantic module.
    The \textit{syntactic module} removes English stop words and collects unigrams, bigrams, and trigrams. Then, for each n-gram, it computes the Levenshtein similarity with the labels of the topics in CSO. Finally, it returns all research topics whose labels have a similarity score equal to or higher than a threshold to one of the n-grams. 
    The \textit{semantic module} uses part-of-speech tagging to identify candidate terms composed of a proper combination of nouns and adjectives and maps them to the ontology topics by using a Word2Vec model trained on titles and abstracts of 4,5M English papers in the field of Computer Science from MAG. 
    Then, the module computes a relevance score for each topic in the ontology by considering the number of times the topic was identified within the retrieved words. 
    The CSO classifier combines the outputs of these two modules and enhances the resulting set by including all relevant super-topics according to the \textit{superTopicOf}\footnote{\url{https://cso.kmi.open.ac.uk/schema/cso}} relationship in CSO. For instance, if an article was tagged with the topic \textit{neural networks}, it would also be associate to its super-topics \textit{machine learning}, \textit{artificial intelligence}, and \textit{computer science}. This latter functionality is not used in the extraction stage, since it would not be possible to map super-topics with other entities in the sentence. However, we used a similar process in the last phase of the process (Knowledge Graph Enhancement, see Section 3.5) for generating further triples by exploiting CSO hierarchical relationships.

    \item \emph{OpenIE~\cite{angeli2015leveraging}} provided by the Stanford Core NLP suite. It detects general entities and relations among them. Relations are detected by analyzing clauses (i.e., groups of words that contain at least a subject and a verb) which are built by exploring the parse tree of the input text.  In the first stage, the methodology produces clauses from long sentences which stand on their own syntactically and semantically. For doing so, it uses a multinomial logistic regression classifier to recursively explore the dependency tree of sentences from governor to dependant nodes. Then, it applies logical inferences to capture natural logic within clauses by using semantics dictating contexts. Doing so, OpenIE is able to replace lexical items with something more generic or more specific. Once a set of short entailed clauses is produced, it segments them into its output triples. In our approach we keep triples where entities match those found by the Extractor Framework and the CSO Classifier, so that we caught only those triples that refer to entities of the target domain.

    \item \emph{The Stanford Core NLP PoS tagger}\footnote{\url{https://nlp.stanford.edu/software/tagger.shtml}} which extracts predicates between the entities identified by the Extractor Framework and the CSO Classifier. More specifically, for each sentence $s_i$ it detects all verbs $V = \{v_0, \ldots, v_k\}$ between each pair of entities $(e_m, e_n)$ of that sentence and generates triples in the form \textless $e_m$, $v$, $e_n$ \textgreater\,\,    where $v \in V$. 
  
    Our goal was to detect the most used verbs between two entities at the cost of producing some noisy relations.
    Indeed, this approach is able to return several additional relationships that were missed by the other tools.
    In the following sections we describe how we handled and validate these triples in order to reduce the noise.
   
\end{itemize}
 
 
We processed each sentence from all the abstracts and used the tools and methods above to assign to each sentence $s_i$ a list of entities $E_i$ and a list of triples $R_i$. 

 First, we run the extractor framework to extract both entities $E_i$ and triples $R_i$. Secondly, we used the CSO Classifier to extract all Computer Science topics, further expanding $E_i$. Thirdly, we processed each sentence $s_i$ with OpenIE, and retrieved all the triples composed by subject, verb, and object in which both subject and object matched the entities resulting from the previous steps. 
 Finally, for each sentence $s_i$ we took all the verbs within two entities through the PoS Tagger, yielding $R_i$ thoroughly expanded.

\subsection{Entities Manager}

During the extraction process it might happen that different entities in $E_i$ may actually refer to the same concept with alternative forms, or may represent too generic concepts that do not carry meaningful information.
In this section, we briefly describe which steps we have performed by the Entities Refiner and Entities Mapper modules in order to address these issues. 

\subsubsection{Entities Refiner Module}
Many of the entities resulting from previous steps can be noisy, ambiguous, and too generic. 

For example, entities like \virgolette{approach} and \virgolette{method} are too abstract and thus not very useful for our purpose. Their presence simply add noise to the the knowledge graph.

The goal of this module is to preprocess the entities, merging alternative labels, discarding ambiguous and generic entities, and splitting the ones that include compound expressions. 

\noindent\textbf{Cleaning up entities.}
First, we removed punctuation (e.g., dots and apostrophes) and stop-words (e.g., pronouns) from entities. We also removed some words that might be mixed up (e.g., \textit{it} might be the pronoun \textit{it} or the acronym of \textit{information technology}) by using a blacklist.

\noindent\textbf{Splitting entities.}
Some entities actually contained multiple compound expressions, e.g., \textit{Machine Learning and Data Mining}. Therefore, we split entities that contain the conjunction \textit{and}. Referring to our example, we obtained the two entities \textit{Machine Learning} and \textit{Data Mining}.

\noindent\textbf{Handling Acronyms.}
Acronyms are usually defined, appearing the first time near their  extended form (e.g., \textit{Web Ontology Language (OWL)}) and then by themselves in the rest of the abstract (e.g., \textit{CSO}). In order to map acronyms with their extended form in a specific abstract we use a regular expression. We then substituted every acronym (e.g., \textit{OWL}) in the abstract with their extended form (e.g., \textit{Web Ontology Language}). Since acronyms can be ambiguous, we perform this operation only on entities from the same abstract.

\noindent\textbf{Detection of Generic Entities.} 
Entities might be too generic for the purpose to describe the knowledge of a domain (e.g., \textit{content}, \textit{time}, \textit{study}, \textit{article}, \textit{input}, and so on). 
We discard these kind of entities by applying a frequency-based filter which compares their frequency in three sets of documents:
\begin{itemize}
    \item the set of publications of the Semantic Web.
    \item a set of the same size covering \textit{Computer Science} domain, but not \textit{Semantic Web}.
    \item a set of the same size containing papers from various domains, but not about \textit{Semantic Web} nor the \textit{Computer Science}.
\end{itemize}

For each entity $e$, we computed the number of times it appeared in the above datasets, so that we had three different counts $c^{'}_e, c^{''}_e, c^{'''}_e$. We normalized the counts by dividing them with the number of words of the set where they were computed. Then we computed the ratios $r^{'}_e = \frac{c^{'}_e}{c^{''}_e}$ and $r^{''}_e = \frac{c^{'}_e}{c^{'''}_e}$. If the ratio $r^{'}_e$ met a threshold $t^{'}_e = 2$, and the ratio $r^{''}_e$ met a threshold $t^{''}_e = 10$, the entity $e$ was included in the graph. Thresholds were empirically defined by manually evaluating which entities were saved/discarded.

In addition, we automatically preserved all the entities within a whitelist that includes CSO topics and the author's keywords of all the papers in the input dataset.

\subsubsection{Entities Merger Module}
We merge entities with the same meaning by using both a lemmatizer and the CSO ontology. Singular and plural forms are combined by using the Lemmatizer available in the SpaCy\footnote{\url{https://spacy.io}} library. Then we exploited the alternative labels described by CSO to merge entities that refer to the same research topic (e.g., "ontology alignement" and "ontology matching"). More specifically, given an entity $e \in E$ that is known by CSO, let $A_{e} = \{e_0, ..., e_{k-1}\}$ be the set of alternatives of $e$ in CSO. The module first finds the longest label $e_{longest} \in A_e$, then $e$ is replaced by $e_{longest}$. The same process is repeated for each entity $f \in E$.

\subsection{Relations Manager}
This step aims at (i) finding the best relation predicate for each pair of entities $e_i, e_j$ where a relation exists (each element in $R$), and (ii) mapping all the relations within a table we have defined.

\subsubsection{Best Relation Finder Module}\label{best-relation}
Here, the set of triples $R$ presents three different types of triples: those extracted by the Extractor Framework, let us say $R_{EF}$, those coming from OpenIE, let us say $R_{OIE}$, and those detected with the PoS tagger, called $R_{PoS}$. We performed the following operations on these sets:
\begin{itemize} 
    \item On the set of triples in $R_{EF}$ we acted as follows. Given a pair of entities $(e_p, e_q)$ in $R_{EF}$, we merged into a list $L_r$ all relations' labels $r_i$ such that  $(e_p, r_i, e_q) \in R_{EF}$. Then we chose the most frequent relation $r_{most\_frequent} \in L_r$, and built a single triple $(e_p, r_{most\_frequent}, e_q)$. Triples so built formed the set $T_{EF}$. Clearly, the size of the set $T_{EF}$ is lower than the size of the set $R_{EF}$.
    
    \item On the set $R_{OIE}$ we performed a deeper merging operation. Similarly to the work performed on $R_{EF}$, given a pair of entities $(e_p, e_q)$ in $R_{OIE}$, we first merged into a list $L_r$ all relations' labels $r_i$ such that  $(e_p, r_i, e_q) \in R_{OIE}$. In $R_{OIE}$ all triples have a verb as relation predicate. Hence, we assigned each $r_i$ to its word embedding $w_i$ from the word embeddings built on the MAG dataset, yielding the list $L_w$. With the word embeddings in $L_w$ an averaged word embedding $w_{avg}$ was built. Then, the relation $r_i$ with the word embedding $w_i$ nearest to $w_{avg}$ according to the cosine similarity was chosen as final relation for the pair $(e_p,e_q)$, yielding the triple $(e_p,w_i,e_q)$. The same procedure  was also applied on $R_{PoS}$. The execution of this procedure on $R_{OIE}$ and $R_{PoS}$ yielded the sets $T_{OIE}$ and $T_{PoS}$, respectively.

    \item { Finally, for the sets $T_{EF}$, $T_{OIE}$, and $T_{PoS}$,   
    we saved for each triple $(e_p, r_i, e_q)$ the number of papers where the pair of entities $(e_p,e_q)$ appeared. We refer to this number as the \textit{support} of triples.}
\end{itemize}

\subsubsection{Mapper Module}
From the previous step, a large number of verb relations resulted. However, the majority of relations have a common meaning with others, i.e., many relations were represented by synonyms. For example, the relations \textit{uses}, \textit{utilizes}, \textit{adopts}, and \textit{employs} may be used to express the same concept within a triple with only a slight change in meaning. Within our triples set, there were a good number of triples that represented the same information such as \textit{\textless ontology alignment, uses, ontology\textgreater} and \textit{\textless ontology alignment, utilizes, ontology\textgreater}. 
Hence, in order to reduce the number of redundant relations, 
we built a map $M : verb\_relation \rightarrow verb\_relation_{representative}$ 
where semantically similar relations  were mapped to a single label 
(e.g., the relations \textit{uses}, \textit{utilizes}, \textit{adopts}, \textit{employs} were mapped to the same representative relation \textit{uses}). In order to do this, we first retrieved all word embeddings that represented verb relations from sets $T_{OIE}$ and $T_{PoS}$. The rationale behind this is that word embeddings represent semantic and syntactic properties of the words and, therefore, verb relations with similar word embeddings have similar semantics and meaning. Then, we used the hierarchical clustering algorithm provided by the SciKit-learn library\footnote{\url{https://scikit-learn.org}}, which uses $(1 - $ cosine similarity) as distance to group together similar verb relations. 
The cosine similarity quantifies the angle between two vectors. Its formula applied on two vectors $v_1$ and $v_2$ can be observed in (\ref{cosine-formula}).  
\begin{equation}\label{cosine-formula}
Cosine\_similarity(v_1, v_2) = \frac{v_1 \cdot v_2}{\parallel v_1 \parallel \cdot \parallel v_2 \parallel}
\end{equation}
The resulting clustering dendrogram was cut by an empirically determined threshold of the averaged Silhoutte-width $=0.65$.
Values of Silhouette width range from $-1$ to $1$. When the value is closer to 1, it means that the clusters are well separated; when the value is closer to 0, it might be difficult to detect the decision boundary; when the value is closer to -1, it means that elements assigned to a cluster might have been erroneously assigned. Its formula is computed as shown in~(\ref{silhouette}), where given a cluster $c$, $w(c)$ represents the average dissimilarity of elements in $c$, and $o(c)$ is the lowest average dissimilarity of elements of $c$ to any other cluster.
\begin{equation}\label{silhouette}
s(c) = \frac{o(c) - w(c)}{max\{o(c),w(c)\}}
   \end{equation}
Subsequently, we manually revised the clusters and built the map $M$, where each verb relations of each cluster was mapped to the representative relation identified by the cluster centroid. For example, each verb relation of the cluster \{\textit{builds}, \textit{creates}, \textit{produces}, \textit{develops}, \textit{makes}, \textit{constructs}, etc.\} was mapped to the centroid \textit{produces}. Finally, the relations used in the set $T_{EF}$ were manually integrated within the map $M$. All triples from the union of $T_{EF}$, $T_{OIE}$, and $T_{PoS}$ were mapped by using $M$ (e.g., the triple  \textit{\textless knowledge construction, creates, ontology integration platform\textgreater} was transformed in \textit{\textless knowledge construction, produces, ontology integration platform\textgreater}), so that a well-defined set of relations was used within our final resulting triples.

\subsection{Triples Selection}
In this section the method we employed to choose only certain triples is presented. We also define what we mean with the words \textit{valid} and \textit{consistent} associated to our triples in order to build the scientific knowledge graph. 

\subsubsection{Valid Triples}
For the purpose of including meaningful triples within our knowledge graph, we first define a smaller knowledge graph composed of "valid" triples. These can be defined in different way according to the performance of the tools in the first step and the number of papers supporting a certain triples.

In the current prototype we define as valid the following triples:  
\begin{itemize}

    \item 
    We consider \textit{valid} all the triples  obtained by the Extractor Framework ($T_{EF}$) and the  OpenIE tool ($T_{OIE}$).
    \item All triples associated with at least 10 papers (indicating a fair consensus). Therefore, we consider \textit{valid}  the triples that were detected by the PoS tagger associated with at least 10 papers. We refer to this set as $T'_{PoS}$ such that $T'_{PoS} \subseteq T_{PoS}$.

\end{itemize}

The union of $T_{TF}$, $T_{OIE}$, and $T'_{PoS}$ composed the set of all valid triples $T_{valid}$.

\subsubsection{Consistent Triples}
\label{consistenttriples}

The set of triples not in $T_{valid}$, that we label $T_{invalid}$, 
may still include several good triples that were not associated to sizable number of papers. More specifically, consensus of the community about scientific claims is built over time and, hence, new discoveries might not have a high support. However, these triples are still important since they can suggest the ways to go along to rapidly explore the most recent research trends. 
We thus use the triples in $T_{valid}$ as examples to learn which triples are consistent with the valid ones and could still be included in the final outcome. Specifically, we trained a classifier $\gamma : P \rightarrow L$ where $P$ is a set of pair of entities blue in $T_{valid}$ and $L$ is the set of relations used in $M$ (e.g., \textit{uses}, \textit{provides}, \textit{supports}, \textit{improves}), with the aim of comparing the actual relation with the one returned by the classifier. The intuition is that a triple consistent with $T_{valid}$ would have its relation correctly guessed by the classifier.
In order to do so, we performed the following steps:

\begin{enumerate}
    \item We generated  word embeddings of size $300$ by processing with the Word2vec algorithm~\cite{mikolov2013efficient, mikolov2013distributed} all the input abstracts. For multi-word entities we replaced white spaces with underscore characters within our abstracts texts (e.g., the entity \textit{semantic web} becomes \textit{semantic\_web}).
    
    \item We trained a Multi-Perceptron Classifier (MLP) to return the  relation between a couple of entities. We used the concatenation of the embeddings of subject and object entities as input and the relation as output.
    
    \item The validation step was performed by applying the classifier on all the triples $(e_p, r, e_q)$ in $T_{invalid}$ and comparing the actual relation $r$ with the relation returned by the classifier $r'$. If $r = r'$ then the triple $(e_p, r, e_q)$ was considered valid and added to $T_{valid}$. Otherwise we computed the cosine similarity $cos\_sim$ and the \textit{Wu-Palmer}\footnote{\url{http://www.nltk.org/howto/wordnet.html}} similarity between the embedding of $r$ and $r'$. If the average between $cos\_sim$ and  $wup\_sim$ was higher than a threshold $t$ (empirically set at $0.5$) then the triple $(e_p, r, e_q)$ was considered valid and added to $T_{valid}$.
\end{enumerate}

\subsection{Knowledge Graph Enhancement}

In order to increase the amount of the resulting information, we added to the produced knowledge graph the additional triples that could be inferred by exploiting the hierarchical relations in CSO. More precisely, given a triple ($e_2$, $r$, $e_1$), if in CSO the entity $e_3$ is \textit{superTopicOf} of the entity $e_1$ and there is no triple involving  $e_2$ and $e_3$, we also infer the triple ($e_2$, $r$, $e_3$). For instance, given the triple \textit{\textless nlp systems, uses,	named-entity recognition \textgreater},  if \textit{artificial intelligence} is \textit{superTopicOf} \textit{named-entity recognition},  we can infer the triple \textit{\textless nlp systems, uses,	artificial intelligence \textgreater}.

This last step was performed by the CSO Triples Integrator module in the pipeline. Finally, the triples are converted to RDF and returned.

\begin{table}[]
\caption{Examples of triples that our pipeline detects. In \textit{Italic} some examples of triples that were discarded by our pipeline.}\label{relationships-example}
\begin{center}
\begin{tabular}{c@{\quad}|@{\quad}c@{\quad}|@{\quad}c}
	\hline\rule{0pt}{12pt}
     \textbf{Subject Entity} & \textbf{Relation} & \textbf{Object Entity}\\
     \hline\rule{0pt}{12pt}
       semantic web technologies & supports & contextual information \\
       semantic relationship & defines & ontologies\\
       structural index & uses & structural graph information\\
       thesaurus & hyponymy-of/is & knowledge organization system\\
       web page classification &	uses & text of web page \\
       question answering systems &	uses & semantic relation interpreter\\
       \textit{context models} & \textit{proposes} & \textit{web ontology language}\\
       \textit{data exchange} & \textit{queries} & \textit{web ontology language}\\
       \textit{domain-specific ontologies} & \textit{executes} & \textit{semantic search engines}\\
        \textit{fuzzy logics} & \textit{maintains} & \textit{semantic descriptions}\\
        \textit{learning objects} & \textit{learns} & \textit{semantic web services}\\
        \textit{resource description framework (rdf)} & \textit{uses}  & \textit{digital libraries} \\
    \hline
\end{tabular}
\end{center}

\end{table}

\section{Results and Discussion}\label{scholarly-result}
This section details the scientific knowledge graph we have produced and shows how we have validated it.

\subsection{The Semantic Knowledge Graph}
Here we report the result of our framework, focusing on the  \textit{Semantic Web} domain.

We used an input dataset composed by 
$26,827$ abstracts of scientific publications about this domain that was retrieved by selecting publications from the Microsoft Academic Graph dataset\footnote{\url{https://www.microsoft.com/en-us/research/project/microsoft-academic-graph}}. It is a knowledge graph related to the scholarly domain that describes more than $200$ million scientific publications through metadata such as title, abstract texts, authors, venue, field of study and so on.  For our purpose we considered only abstracts that were classified under Semantic Web by the CSO Classifier~\cite{salatino2019cso}. This dataset has also been used for exploring the relationship between Academia and Industry by Angioni et al. ~\cite{angioni2019integrating}. 

A few examples of retrieved triples as well as of triples that were discarded by our pipeline can be seen in Table~\ref{relationships-example}.

The resulting knowledge graph includes 
$109,105$ triples: $87,030$ from the Extractor Framework ($T_{EF}$), $8,060$ from OpenIE ($T_{OIE}$), and $14,015$ from the PoS tagger method and classifier ($T'_{PoS}$ + \textit{Cons. Triples}).

However, the raw number of triples extracted by each method can be misleading. 
In fact, some triples are supported by a large number of papers, suggesting a large consensus of the scientific community and more in general a claim that can easily be trusted, while some other appear in one  or very few papers

\begin{figure}[!t]
    \centering
    \includegraphics[width=1.0\linewidth]{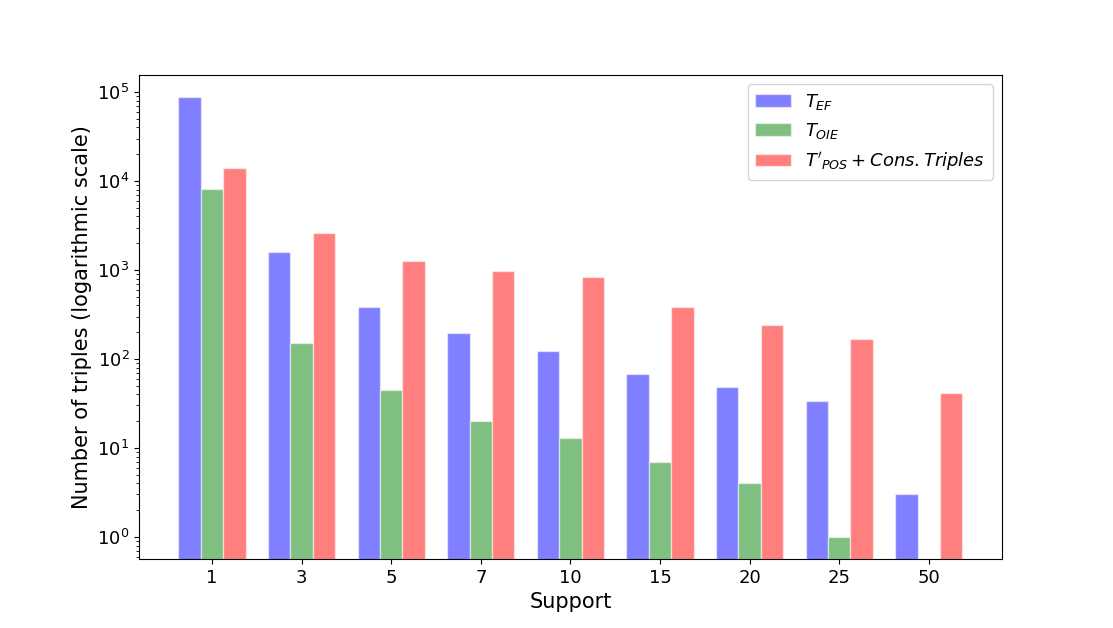}
    \caption{Comparison of the distribution of the support of the three methods.}
    \label{fig:grouped_plot_triples}
\end{figure}

Figure~\ref{fig:grouped_plot_triples} reports the distribution of the support of the triples produced by $T_{EF}$, $T_{OIE}$ and $T'_{PoS}$ + \textit{Cons. Triples}.

While $T_{EF}$ produces the most sizable part of those triples, most of them have a very low support. In fact, $80,030$ of them are supported by a single paper and $1,580$ by only three papers. They may thus contain claims that did not reach yet a consensus in the community.  For all the other support values, the set $T'_{PoS}$ + \textit{Cons. Triples} has a higher number of triples than $T_{EF}$ and $T_{OIE}$ and, hence, it is possible to assume that $T'_{PoS}$ triples may be more in accordance within the community of \textit{Semantic Web}.
For instance, if we take in consideration only the triples whose support is equal or greater than $5$, only $393$ triples are provided by the set $T_{EF}$, $45$ by $T_{OIE}$ and, $1,268$ by $T'_{PoS}$ + \textit{Cons. Triples}.  It is also worth to note that when the support is very high (e.g., equal or greater than $50$) there are not triples provided by the set $T_{OIE}$, and few triples provided by $T_{EF}$. This still stresses the fact that those triples might not express valuable knowledge or have consensus within the \textit{Semantic Web} community.

\subsection{Gold Standard Creation}

We first used several different approaches to generate triples from the $26,827$ abstracts described in the previous section. Specifically, we applied on this dataset: 1) $T_{EF}$ (i.e., the Extractor Framework), 2) $T_{OIE}$ (i.e., OpenIE), 3) $T'_{PoS} $ (considering only the triples with support $\geq 10$), and $T'_{PoS} $ + \textit{Cons. Triples}.

\begin{figure}[!t]
    \centering
    \includegraphics[width=0.7\linewidth]{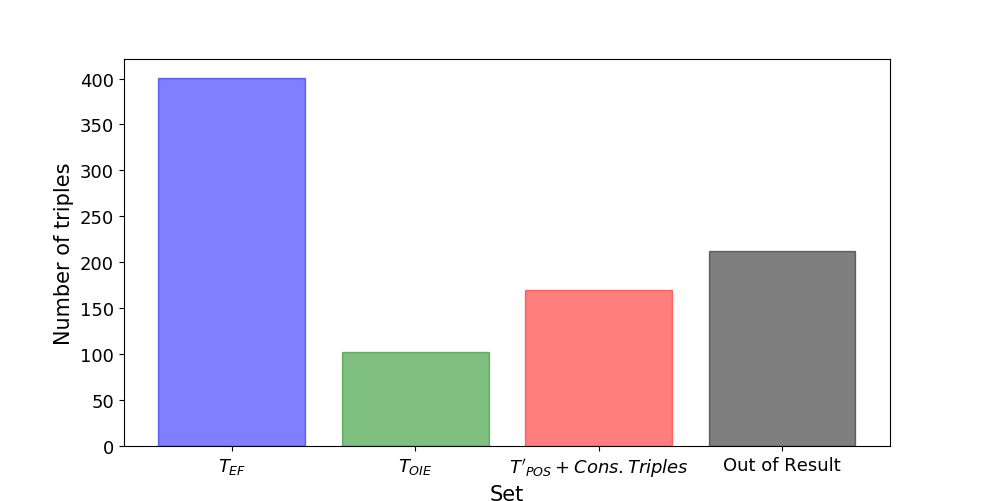}
    \caption{Distribution of triples within the gold standard.}
    \label{fig:gs_distr}
\end{figure}

The resulting set of $109,105$ triples would be unfeasible to manually annotate, since it is very large and includes terms related to very different areas of expertise. We thus focused only on $818$ triples which contain (as subject or object) at least one of the 24 sub-topics\footnote{There exist 24 sub-topics of Semantic Web within the CSO ontology.} of Semantic Web and at least another topic in the CSO ontology. This set 
contains 401 triples from  $T_{EF}$, 102 from $T_{OIE}$, 60 triples from  $T'_{PoS} $ and 110 relevant \textit{Cons. Triples}. In order to measure the recall, we also added 212 triples that were discarded by the framework pipeline. The reader notices that the total number of triples (818) is slightly less than the sum of various sets (401+102+60+110+212) because some triples have been derived by more than one tool. The triples distribution of the gold standard can be observed in Figure~\ref{fig:gs_distr}.

We recruited five researchers in the field of Semantic Web and asked them to annotate each triple either as \textit{true} or \textit{false}.
In order to do so, they assessed each triple according to their expertise of the field. They were also allowed to search concepts on the web and in the literature when they were not familiar with a specific entity.
The averaged agreement between experts was $0.747 \pm 0.036$, which indicates a high inter-rater agreement. 
We then created the gold standard using the majority rule approach. Specifically, if a triple was considered relevant by at least three annotators, it was labeled as true, otherwise as false.

The purpose of this gold standard is twofold. First, it allows us to evaluate the proposed pipeline to extract triples from scholarly data and, second, it provides  a  resource  which  will  facilitate  further  evaluations.

\subsection{Precision, Recall, F-measure Analysis}
For evaluating our methodology, we performed a precision, recall, F-measure analysis considering various combinations of relations sources. Measures are computed as shown by equations~(\ref{precision-eq}), (\ref{recall-eq}), and~(\ref{fmeasure-eq}).

\begin{equation}\label{precision-eq}
P = \frac{TP}{TP + FP}
\end{equation}

\begin{equation}\label{recall-eq}
R = \frac{TP}{TP + FN}
\end{equation}

\begin{equation}\label{fmeasure-eq}
F = 2 \cdot \frac{P \cdot R}{P + R}
\end{equation}

In equations~(\ref{precision-eq}) and~(\ref{recall-eq}) $TP$ (true positive) indicates the number of triples labelled as \textit{good} and returned by our pipeline, $FN$ (false negative) is the number of triples that were labelled as \textit{good} but not returned by our pipeline, and $FP$ (false positive) is the number of triples that have been erroneously returned by our pipeline (i.e., triples were labelled as \textit{bad} in the gold standard but our pipeline picked up them as \textit{good} triples). The F-measure is computed as the harmonic  mean of~(\ref{precision-eq}) and~(\ref{recall-eq}) as shown in equation~(\ref{fmeasure-eq}).

We tested eight alternative approaches:
\begin{itemize} 
    \item The Extractor Framework from Luan Yi et al.~\cite{luan2018multi} (\textbf{EF}) described in section~\ref{extraction_entities_relations}.
    \item OpenIE, from Angeli et al.~\cite{angeli2015leveraging}  (\textbf{OpenIE}) described in section~\ref{extraction_entities_relations}.
    \item the Stanford Core NLP PoS tagger described in section~\ref{extraction_entities_relations}, after merging the relevant triples as described in section~\ref{best-relation}
    (\textbf{$T'_{PoS}$}). We considered only the triples with support $\geq 10$.
    \item The previous approach enriched by consistent triples as described in section \ref{consistenttriples} 
    (\textbf{$T'_{PoS}$ + Cons. Triples}).
    \item The combination of EF and OpenIE (\textbf{EF + OpenIE}).
    \item The combination of EF and $T'_{PoS}$ + Cons. Triples (\textbf{EF + $T'_{PoS}$ + Cons. Triples}).
    \item The combination of OpenIE and $T'_{PoS}$ + Cons. Triples (\textbf{OpenIE + $T'_{PoS}$ + Cons. Triples}).
    \item The final framework that integrates all the previous methods (\textbf{OpenIE + EF + $T'_{PoS}$ + Cons. Triples}).
\end{itemize}

Table~\ref{relationships-evaluation} reports  precision, recall, and F-measure of all the methods. 

EF obtains an high level of precision (84.3\%), but a recall of only 54.4\%. OpenIE and $T'_{PoS}$ shows a slightly lower level of precision and an even lower recall. $T'_{PoS}$ + Cons. Triples obtains the best precision of all the methods (84.7\%), highlighting the advantages of using a classifier for selecting consistent triples. Overall, all these  basic methods produce triples with good precision, but suffer in term of recall. 

Combining them together generally raises the recall without paying too much in term of precision. EF + OpenIE yields a F-measure of 72.8\% with a recall of 65.1\% and EF + $T'_{PoS}$ + Cons. Triples a F-measure of 77.1\% with a recall of 71.6\%. The final version of our framework, which combines all the previous methods, obtains the best recall (80.2\%) and F-measure (81.2\%) and yields also a fairly good precision (78.7\%). This seems to confirm the hypothesis that an hybrid framework combining supervised and unsupervised methods would produce the most comprehensive set of triples and the best performance overall.

\begin{table}[!b]
\caption{Precision, Recall, and F-measure of each method adopted to extract triples. To note that the last row identified the triples extracted using the full pipeline.}
\begin{center}
\begin{tabular}{c@{\quad}|@{\quad}c@{\quad}|@{\quad}c@{\quad}|@{\quad}c}
	\hline\rule{0pt}{12pt}
     \textbf{Triples identified by} & \textbf{Precision} & \textbf{Recall} & \textbf{F-measure}\\
     \hline\rule{0pt}{12pt}
        EF & 0.8429 & 0.5443 & 0.6615\\
        OpenIE & 0.7843 & 0.1288 & 0.2213\\
        $T'_{PoS}$ & 0.8000 & 0.0773 & 0.1410\\
        $T'_{PoS}$ + Cons. Triples & \textbf{0.8471} & 0.2319 & 0.3641\\
        EF + OpenIE & 0.8279 & 0.6506 & 0.7286\\
        EF + $T'_{PoS}$ + Cons. Triples & 0.8349 & 0.7166 & 0.7712\\
        OpenIE + $T'_{PoS}$ + Cons. Triples & 0.8145 & 0.3253 & 0.4649\\
        OpenIE + EF + $T'_{PoS}$ + Cons. Triples& 0.7871 & \textbf{0.8019} &\textbf{0.8117}\\
    \hline
\end{tabular}
\end{center}
\label{relationships-evaluation}
\end{table}

\subsection{Examples and considerations about the Scientific Knowledge Graph}

In this section, we show some sample of the triples extracted for the Semantic Web Knowledge Graph and discuss benefits and limitations of our output.

Table~\ref{triples-example_1} shows a selection of the triples about the research topic \textit{ontology alignment}, ranked by \textit{support}.
It is easy to see that many of these triples define the fundamental characteristics of \textit{ontology alignment}. The topic is contextualized (via "skos:broader" relations) within the areas of \textit{semantic web technologies} and \textit{information integration}. \textit{Ontology alignment} is defined as an entity that uses \textit{ontologies}, selects \textit{semantic correspondences}, and supports \textit{semantic interoperability}.

Several other triples add further  details, such as that \textit{ontology alignment} finds \textit{semantically related entities}, adopts \textit{semantic similarity measures}, and limits the need for \textit{human intervention}. Naturally, the representation also suffers from some issues that we plan to address in future work. For instance, the triples \textit{\textless ontology alignment, selects, mapping\textgreater} and \textit{\textless ontology alignment, supports, semantic relations\textgreater} appear too ambiguous. This may be either a limitation of our vocabulary of relations or an issue in the methodology used for merging together the triples from the PoS tagger.  Similarly, in \textit{\textless ontology alignment, produces,	semantic web application\textgreater} the predicate does not appear to be correct, maybe "support" would be a better choice in this case. We thus plan to work further on our approach for merging triples and select the best predicate between two entities. 

The triple \textit{\textless ontology alignment, produces,	semantic web application \textgreater} shows another typical issue. In the knowledge graph we have both "distributed and heterogeneous ontology" and "heterogeneous ontology" but no link between the two. In the future we need to be able to detect that "distributed and heterogeneous ontology" is actually a sub-concept of "heterogeneous ontology".

\begin{table}[]
\caption{Examples of triples from the Semantic Web Knowledge Graph.}
\begin{center}
\begin{tabular}{c@{\quad}|@{\quad}c@{\quad}|@{\quad}c |@{\quad}c}
	\hline\rule{0pt}{12pt}
     \textbf{Subject Entity} & \textbf{Relation} & \textbf{Object Entity} & \textbf{Support}\\
     \hline\rule{0pt}{12pt}
       ontology alignment & uses & ontologies & 194 \\
       ontology alignment & skos:broader & semantic web technologies & 65 \\
       ontology alignment & selects & semantic correspondence & 45 \\
       ontology alignment & supports & semantic interoperability & 34 \\
       ontology alignment & maintains & heterogeneous ontology & 25 \\
       ontology alignment & selects & mapping & 21 \\
       ontology alignment & selects & semantically related entity & 19 \\
       ontology alignment & supports & semantic relation & 17 \\
       ontology alignment & produces & semantic web application & 14 \\
       ontology alignment & combines & concept similarity & 13 \\       
       ontology alignment & supports & semantic heterogeneity problem & 13 \\
       ontology alignment & limits & human intervention & 12 \\
       ontology alignment & executes & semantic similarity measures & 12 \\
       ontology alignment & produces & ontology mapping method & 11 \\
       ontology alignment & provides & distributed and heterogeneous ontology & 10 \\
       ontology alignment & skos:broader & information integration & 10 \\
       ontology alignment & uses & mapping system & 10 \\
       ontology alignment & provides & matching technique & 10 \\
    \hline
\end{tabular}
\end{center}
\label{triples-example_1}
\end{table}

\begin{figure}[!tb]
    \centering
    \includegraphics[width=\linewidth]{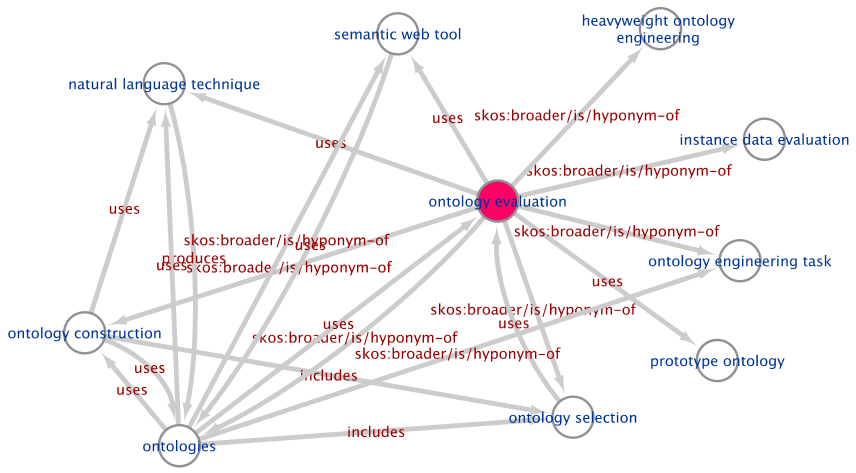}
    \caption{The subgraph of the entity "ontology evaluation" with related relationships in our Scientific Knowledge Graph within the Semantic Web domain.}
    \label{fig:ontology_evaluation}
\end{figure}

 Figure~\ref{fig:ontology_evaluation} shows a graphical representation of the research topic \textit{ontology evaluation}. 
 It is interesting to notice how this representation is also fairly interpretable by human users. Some examples about the information that can be derived includes:
 
\begin{itemize}
    \item \textit{ontology evaluation} uses \textit{natural language techniques}. It suggests that there might be tools or methodologies that exploit textual resources written in natural language that have been involved in ontologies evaluation.
    \item \textit{ontology evaluation} is hyponym of the entity \textit{ontology construction} indicating that a specialized task within \textit{ontology construction}  involves the evaluation of the produced ontologies.
    \item \textit{ontology evaluation} is hyponym of \textit{instance data evaluation} which shows in which more general task the \textit{ontology evaluation} falls.
\end{itemize}

\begin{figure}[!t]
    \centering
    \includegraphics[width=0.8\linewidth]{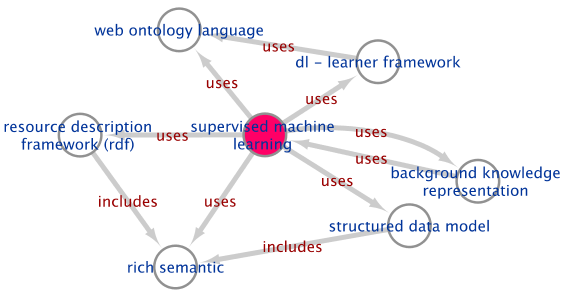}
    \caption{The subgraph of the entity "supervised machine learning" with related relationships in the produced Scientific Knowledge Graph within the Semantic Web domain.}
    \label{fig:supervised_machine_learning}
\end{figure}

Finally, it is also interesting to consider an entity that is not so much represented in the input dataset. Figure~\ref{fig:supervised_machine_learning} shows the subgraph of the entity \textit{supervised machine learning}. This representation is useful to highlight which topics and kind of resources are employed by \textit{supervised machine learning} within the \textit{Semantic Web} domain. As an example, it is easy to see that this entity uses both \textit{structured data model} and \textit{rich semantics}, and how these two entities are related as well. In the example, only two types of relations appear (i.e., \textit{uses} and \textit{includes}). They seem too generic, in fact, it is not clear how  \textit{supervised machine learning} adopts the other linked entities. This can indicate that our  taxonomy of predicates may be too general and we may have to adopt a more fine grained representation in future work.

Overall, the knowledge graph seems to contain triples of good quality that well represent the main characteristics of research entities within the context of the input dataset. We thus believe that this version may already be used for enhancing the representation of research items and supporting users in understanding and navigating research outcomes.

Specifically, we see four main applications of the knowledge graph. The first regard intelligent systems for navigating research publications, such as Open Knowledge Maps\footnote{https://openknowledgemaps.org/}, which could further characterize entities according to their types and relationships and thus interlinking articles according to a variety of new facets and generating more semantically consistent clusters of articles.  The knowledge graph should also be of interest for the growing area of graph embeddings. Indeed, we received several queries by research groups interested in running methods for producing graph embeddings on our output in order to generate a representation of the research entities that could be easily fed to machine learning algorithms and link detection techniques. 
Systems for recommending research papers (e.g., Mendeley\footnote{https://www.mendeley.com/}, CORE\footnote{https://core.ac.uk/}) could also take advantage of this knowledge base for improving and explaining their suggestions according to the entities in the articles. Finally, trend detection systems (e.g., Augur \cite{salatino2018}, ResearchFlow \cite{salatino2020researchflow}), which typically identify entities of interest from a vocabulary or a domain taxonomy and monitor them across time, will benefit by having a large knowledge base of well-defined and interlinked research entities.

\subsection{Applicability in other domains}
In this section, we discuss which the current limitations of our pipeline to be used within other domains are, and suggest some developments that are required to capture other domain peculiarities. To start with, the current version of the proposed pipeline exploits both computer science-tuned modules (e.g., the extractor framework and the CSO classifier) and others more general ones that do not depend on the domain (e.g., OpenIE and the PoS Tagger extractors). Moreover, the handling of entities and relations does not depend on the target domain and, therefore, the pipeline is limited to the computer science field only by some tools employed in the extraction phase. More specifically, the extractor framework  was trained on a corpus of computer science scientific papers, and the ontology employed only embraces computer science topics. As a matter of principle, this implies that the current pipeline can be exploited in any computer science sub-field without limitations. To use the pipeline on other domains, the main challenges are the substitution of the Extractor Framework and the Computer Science Ontology. However, this limitation can be easily tackled in many domains where there already exist tools that can be used to parse the domain scientific resources. To name an example, the \textit{SciSpacy}\footnote{\url{https://allenai.github.io/scispacy/}} model can be used to parse scientific text within the biomedical domain to detect the research entities that characterize it. In the same way, most scientific disciplines offer domain ontologies or taxonomies that could be used in alternative to CSO. These includes the Medical Subject Heading (MeSH)\footnote{Medical Subject Heading - \url{https://www.ncbi.nlm.nih.gov/mesh}} and \textit{SNOMED-CT}\footnote{\url{http://www.snomed.org/snomed-ct/five-step-briefing}} in Biology, the  Mathematics Subject Classification (MSC)\footnote{Mathematics Subject Classification - \url{https://mathscinet.ams.org/msc}} in Mathematics, and the Physics Subject Headings (PhySH)\footnote{Physics Subject Headings - \url{https://physh.aps.org/}} in Physics.  
Broadly speaking, the tools we used to detect the first entities and relations can be replaced by existing tools that have been already developed in other domains to capture domain specific information. One more point to be considered is that ontological resources are today being developed for many specific domains and use cases such as the Cultural Heritage domain (e.g, ArCO~\cite{carriero2019arco}), Robotics~\cite{bardaro2019parsing}, Bio-Medicine\footnote{\url{https://bioportal.bioontology.org/ontologies}}, Computer Science (e.g., AIDA~\cite{angioni2020integrating}), and so on. Therefore, the main  efforts might be due to the developments of interfaces to feed our pipeline with new extraction resources output.

\section{Related Work}\label{related}
Many information extraction approaches for harvesting entities and relationships from textual resources can be found in literature.
 
First, entities in textual resources have been detected by applying Part-Of-Speech (PoS) tags. An example is constituted by~\cite{moro2014entity}, where authors provided a graph based approach for Word Sense Disambiguation (WSD) and Entity Linking (EL) named Babelfly. Later, other approaches started to exploit various resources (e.g., context information and existing knowledge graphs) for  developing ensemble methodologies~\cite{martinez2018openie}. Following this idea, we exploited an ensemble of tools to mine scientific publications and get information out of them. Then, we designed and implemented a software pipeline for the purpose of creating a scientific knowledge graph that organizes entities and their relations. 
Relations extraction is not a novel task and has been already addressed in literature in order to connect information coming from different pieces of text. FRED\footnote{\url{http://wit.istc.cnr.it/stlab-tools/fred/}} is a machine reader developed by~\cite{GangemiEtAl2017}  on top of Boxer~\cite{curran2007linguistically}. It links elements to various ontologies in order to represent the content of a text in a RDF representation. Among its features FRED extracts relations between frames, events, concepts and entities. However, integrating its extracted knowledge for specific domain applications still remains an open challenge due to the unpredictable and too generic type of knowledge that is extracted, making difficult the use of its entities and relations for modelling scholarly contents. 
Moreover, FRED only considers a single text at a time and does not consider domain dependent characteristics that different sources may have. Differently from Gangemi et al.~\cite{GangemiEtAl2017}, our approach aims at parsing specific type of textual data and, moreover, at combining information from various textual resources. For this purpose, we combined results of open domain information extraction tools with information related to the scholarly domain. Furthermore, within our approach more scientific papers are parsed in order to come up with knowledge resulting from the synthesis of various pieces of texts that refer to the same topic. 
With our approach the resulting scientific knowledge graph represents the overall knowledge presented within the input scientific publications.

Researchers have already targeted scientific publications as a challenge domain where to extract structured information~\cite{10.1007/978-3-319-24282-8_18,nuimeprn6347}. Furthermore, within the scholarly domain, extraction of relations from scientific papers has recently raised interest within the \textit{SemEval 2017 Task 10: ScienceIE - Extracting Keyphrases and Relations from Scientific Publications}~\cite{augenstein-etal-2017-semeval} and SemEval 2018 Task 7 \textit{Semantic Relation Extraction and Classification in Scientific Papers} challenge~\cite{gabor2018semeval}, where participants had to face the problem of detecting and classifying domain-specific semantic relations. Since then, extraction methodologies for the purpose to build knowledge graphs from scientific papers started to spread in literature~\cite{10.1007/978-3-030-33982-1_5,LABROPOULOU18.13}. For example, Al-Zaidy et al.~\cite{al2018extracting} employed syntactical patterns to detect entities, and defined two types of relations that may exist between two entities (i.e., \textit{hyponymy} and \textit{attributes}) by defining rules on noun phrases.
Another attempt to build scientific knowledge graphs from scholarly data was performed by Yi and colleagues~\cite{luan2018multi}, as an evolution of authors' work at SemEval 2018 Task 7. First, authors proposed a Deep Learning approach to extract entities and relations from scientific literature. Then, they used the retrieved triples for building a knowledge graph on a dataset of $110,000$ papers. Although our work takes inspiration from that, we propose different strategies to address open issues for combining entities and relations. For example, for solving ambiguity issues that regard the various representations of entities, Yi and colleagues~\cite{luan2018multi} considered clusters of co-referenced entities to come up with a representative entity in the cluster. On the contrary, we adopted textual and statistics similarity to solve the ambiguity. Furthermore, they only used a set of predefined relations that might be too generic for the purpose of yielding insights from the research landscape. Within our approach we tried to detect relations that imply an action of an entity toward another one, making our results more precise and fine-grained.

\section{Conclusions}\label{conclusion}
In this paper we designed and developed a pipeline for representing the knowledge of scientific publication into a structured graph that we called scientific knowledge graph. We employed various state-of-the-art NLP tools and machine learning, and provided a workflow to merge their results. Moreover, we integrated the knowledge coming from many scientific publications into a single knowledge graph with the purpose to represent detailed knowledge of the scientific literature about the \textit{Semantic Web} domain. The evaluation proved that this solution is able to automatically produce good quality scientific knowledge graphs and that the integration of different tools yields a better overall performance.

There are a number of limitations that need to be still addressed in future work. In the first instance, the current version does not take full advantage of the semantic characterization of the research entities to verify the resulting triples. For instance, it is currently possible for an entity of kind \textit{Material} to include a entity of kind \textit{Task}, which may be semantically incorrect. For this reason, we plan to develop a more robust semantic framework that could drive the extraction process and discard triples that do not follow specific constraints. For example, we could state that a material could include another material, but not a task or a method. These requirements could be enforced and verified with the use of specific semantic technologies for expressing constraints such as SHACL\footnote{https://www.w3.org/TR/shacl/}. 
A second limitation is that the current prototype can only extract one relationship between two entities. This is not completely realistic since two entities can be linked by many kinds of relationships. 
This could also lead to a higher number of relationships that could suggest different applications or uses of entities, increasing the probability of finding unconsidered issues and solutions within a research field. We intend to explore this possibility in future work. Additionally, we will thoroughly investigate the conjunction construct which might hide rich knowledge about the relationship that frequently occurs between two research entities (e.g., \textit{machine learning} and \textit{data mining}).  We also plan to improve the knowledge graph by considering cross document relations (e.g., citations) to further link our entities, in order to better support tools for scientific inquiry. 
A third limitation regards our ability to recognize synonyms that are not defined in existent knowledge bases, such as CSO. For instance, the current version may still fail to recognize that two quite different strings (e.g., Radial Basis Function Neural Network and RBFNN) actually refer to the same entity. 
We intend to address this issue by computing the semantic similarity between word and graph embeddings representing the entities in order to detect and merge synonyms more effectively.  
A fourth limitation regards the scalability of our pipeline. 
The current implementation presents a few bottlenecks that could make difficult to apply it on very large-scale datasets. First, the Extractor Framework requires a lot of hard disk space. This entails that data must be sampled to be processed. Second, the current pipeline only adopts the Stanford Core NLP server with just one thread, which requires a long time to mine textual resources sentence-by-sentence. 
However, this is not a big issue since it would be possible to run the Stanford Core NLP server in multi-thread mode, speeding up the extraction process.   
An important next step will also be  to perform an extrinsic evaluation of the proposed knowledge base within different tasks. In particular, we would like to assess how AI tasks such as those tackled by recommender systems or graph embeddings creation strategies can benefit from it.

\section*{Acknowledgements}
Danilo Dess\`{i} acknowledges Sardinia Regional Government for the financial support of his PhD scholarship (P.O.R. Sardegna F.S.E. Operational Programme of the Autonomous Region of Sardinia, European Social Fund 2014-2020). This work has also been partially supported by a public grant overseen by the French National Research Agency (ANR) as part of the program “Investissements d’Avenir” (reference: ANR-10-LABX-0083). It contributes to the IdEx Université de Paris - ANR-18-IDEX-0001. We gratefully acknowledge the support of NVIDIA Corporation with the donation of the Titan X Pascal GPU used for this research.

\bibliographystyle{elsarticle-num.bst}
\bibliography{main.bib}

\end{document}